\documentclass[mlmain]{jmlr}

\usepackage{longtable}

\usepackage{booktabs}
\usepackage[load-configurations=version-1]{siunitx} 

\theorembodyfont{\upshape}
\theoremheaderfont{\scshape}
\theorempostheader{:}
\theoremsep{\newline}

\usepackage{subcaption}

\jmlrvolume{}
\firstpageno{1}
\editors{List of editors' names}

\jmlryear{2026}
\jmlrworkshop{Symmetry and Geometry in Neural Representations}

\editor{}
\jmlrproceedings{PMLR}{Proceedings Track}

\title[A Variational Account of Processing Fluency in Face Perception]{
Beauty is in the ELBO of the Beholder:
A Variational Account of Processing Fluency in Face Perception
}

\author[López \& Triesch]{\Name{Francisco M. López}\Email{fm.lopez@unsw.edu.au} \\
\addr
University of New South Wales, Sydney, Australia
\AND
\Name{Jochen Triesch} \Email{triesch@fias.uni-frankfurt.de} \\
\addr
Frankfurt Institute for Advanced Studies, Germany
}

\begin{document}

\maketitle

\begin{abstract}
Facial attractiveness has been linked to statistical regularities such as symmetry and averageness, suggesting that beauty may depend on the ease with which a face is perceived. We empirically test this hypothesis by training variational autoencoders on four face datasets without attractiveness supervision and evaluating their representations on the 597 faces from the Chicago Face Database. We report three main novel findings. First, we identify that human attractiveness ratings closely aligns with the direction defined by the VAE evidence lower bound (ELBO) in rate-distortion space. Second, we find that independently learned latent spaces contain an attractiveness direction that transfers strongly across random initializations and training data. Finally, we observe that attractive faces are more prototypical in both shape and latent space. Our work provides the first explicit connection between classic accounts of aesthetics with learned generative models and empirical support for a variational interpretation of the processing fluency theory of aesthetic pleasure.
\end{abstract}
\begin{keywords}
Variational autoencoders; Generative models; Rate-distortion; Attractiveness
\end{keywords}

\section{Introduction}
\label{sec:introduction}

What makes a face beautiful? Classic accounts in face perception link attractiveness to statistical regularities such as symmetry \citep{rhodes1999average,perrett1999symmetry,bertamini2019symmetry}, averageness \citep{langlois1990attractive,rhodes1999average,trujillo2014beauty,lee2025further,burton2005robust}, sexual dimorphism \citep{hoss2005role,lee2025further,kleisner2024distinctiveness}, and overall prototypicality within a population \citep{ryali2020likely}. These findings suggest that beauty may depend on the ease with which a face is perceived by the observer. The processing fluency theory establishes that stimuli that can be represented or retrieved more easily are typically evaluated more positively \citep{reber1998effects,reber2004processing}. While this fluency heuristic has been extensively studied in behavioral experiments \citep{reber1998effects,reber2004processing,winkielman2006prototypes,trujillo2014beauty,ryali2020likely}, it remains unclear how processing fluency emerges in representation learning and how it can be instantiated in computational models of face perception.

Previous computational accounts have proposed explanations of aesthetic value in terms of the efficiency or predictability of perceptual representations. Efficient coding accounts have suggested that preferences may arise from the adaptation of the perceptual system to the sensory statistics \citep{renoult2016beauty, lopez2024prioritizing,lopez2026efficient}. Similarly, predictive coding approaches emphasize uncertainty reduction as a source of aesthetic preference \citep{frascaroli2023aesthetics,van2023order}. Most closely related to our proposal, recent works have formalized aesthetic value under probabilistic scalar models \citep{brielmann2022computational,brielmann2023modelling,nath2026relating}, with immediate reward determined by the fluency of processing the stimulus and an additional component capturing expected learning progress \citep{schmidhuber2008driven}. These works define the core of computational neuroaesthetics research but rely on overly abstract models that lack the detailed face representation qualities of modern generative models, e.g., \citet{karras2019style,higgins2017beta} (see \citet{phillips2026state} for a recent review).

Addressing this mismatch between theories of neuroaesthetics and performant generative models of faces, we inquire how processing fluency can be modeled in generative models. Variational autoencoders (VAEs) \citep{kingma2013auto,kingma2017variational,higgins2017beta} learn probabilistic representations that balance two competing components in their compressive bottlenecks: preserving enough information about an input to reconstruct it accurately and maintaining a regularized latent representation that remains close to a prior distribution. We postulate that this unsupervised representation learning objective, known as the evidence lower bound (ELBO) \citep{kingma2013auto}, provides a computationally explicit notion of processing fluency. The ease with which a stimulus is represented can be quantified by this trade-off between the reconstruction distortion and the regularization rate.

We investigate this hypothesis using convolutional VAEs trained independently on four face image datasets without attractiveness supervision. We then evaluate all learned representations on the 597 faces from the Chicago Face Database (CFD; \citealt{ma2015chicago}), for which human attractiveness ratings are available. We find that human attractiveness gradients closely match the ELBO in the rate-distortion space across training datasets and demographics. Independently learned latent spaces contain strongly aligned attractiveness directions. Attractive faces are also more prototypical in both shape and latent space and have their facial geometry represented more faithfully. Together, these findings support a variational interpretation of processing fluency in facial attractiveness. Our work provides the first explicit connection between classical accounts of face perception, processing fluency theories of aesthetic pleasure, and modern generative models.

\section{A variational interpretation of processing fluency}
\label{sec:variational_fluency}

The processing fluency theory posits that stimuli that are easier to encode, retrieve, or interpret tend to elicit more positive evaluations from human subjects \citep{reber1998effects,reber2004processing}. In this view, preferences are not determined by the objective physical properties of a stimulus in isolation, but rather by how that stimulus ``fits'' into the observer's perceptual system. The familiarity \citep{duke2014parallel}, prototypicality \citep{winkielman2006prototypes,ryali2020likely}, and predictability \citep{ryali2020likely} of an experience can all increase its fluency and therefore its preference. Consequently, processing fluency emerges from the organization of neural representations around the sensory experiences of each individual subject. Processing fluency has not yet been modeled explicitly in modern face representation models \citep{phillips2026state}. 

Probabilistic generative models provide a natural framework for modeling processing fluency. Perception can be viewed as inference under an internal model of how sensory observations are generated \citep{helholtz1867handbuch,friston2010free}. Intuitively, a probabilistic generative model must encode enough information about a sensory observation to reconstruct it accurately, while keeping the resulting internal code compatible with the distribution of codes learned from observations. Given an observation $x$, the system infers latent causes $z$ that can explain that observation. VAEs make use of this principle explicitly by learning an approximate posterior $q_\phi(z\mid x)$ together with a generative model $p_\theta(x\mid z)$ under a prior $p(z)$. Because the exact marginal likelihood is generally intractable, VAEs minimize the negative evidence lower bound (ELBO) \citep{kingma2013auto} given by:
\begin{equation}
-\mathrm{ELBO}(x)
=
\underbrace{
    \mathbb{E}_{q_\phi(z\mid x)}
    \left[-\log p_\theta(x\mid z)\right]
}_{\text{distortion} \;D(x)}
+ \,
\underbrace{
    \mathrm{KL}
    \left[
        q_\phi(z\mid x)\,\|\,p(z)
    \right]
}_{\text{rate} \; R(x)} \; .
\label{eq:elbo}
\end{equation}

The first term measures the fidelity of the latent representation in terms of how well it explains the sensory input. For a Gaussian decoder with fixed variance, this distortion is proportional to a squared reconstruction error \citep{kingma2017variational}. The second term is the regularization rate of the latent space. It measures how far the posterior deviates from the prior, quantified as a Kullback-Leibler (KL) divergence. More generally, the $\beta$-VAE \citep{higgins2017beta} objective generalizes the ELBO to the objective $\mathcal{L}_\beta(x)=D(x)+\beta R(x)$, where the parameter $\beta$ acts as a rate-distortion trade-off.

This decomposition suggests a concrete variational interpretation of processing fluency. A stimulus is represented with ease when it can be explained with high fidelity (low distortion) and at a low representational cost (rate). The ELBO combines these two desiderata into a single quantity and thus provides a scalar measure for processing fluency. Under this interpretation, we predict that aesthetic pleasure stems from a high ELBO (low distortion and low rate) induced by the learned generative representation.

\begin{figure}[!t]
\floatconts
  {fig:reconstructions}
  {\caption{ \label{fig:examples}
    Examples of original face images and their reconstructions from an independently trained VAE, all from the Latino-Male category of the Chicago Face Database (CFD), ordered by increasing human attractiveness rating. Numbers above the original faces indicate within-group standardized human attractiveness ratings. Numbers above the reconstructions indicate within-group
    standardized ELBO ratings. Examples for all other models and demographic groups are provided in Appendix \ref{app:reconstructions}.
  }}
  {%
    \subfigure[Original face images from CFD \citep{ma2015chicago}]{%
      \includegraphics[width=\linewidth]{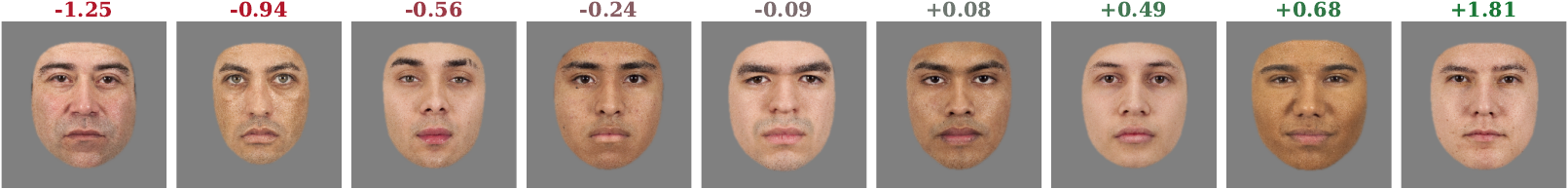}}%

    \vspace{.25em}

    \subfigure[Reconstructions from VAE trained on FairFace \citep{karkkainen2021fairface}]{%
      \includegraphics[width=\linewidth]{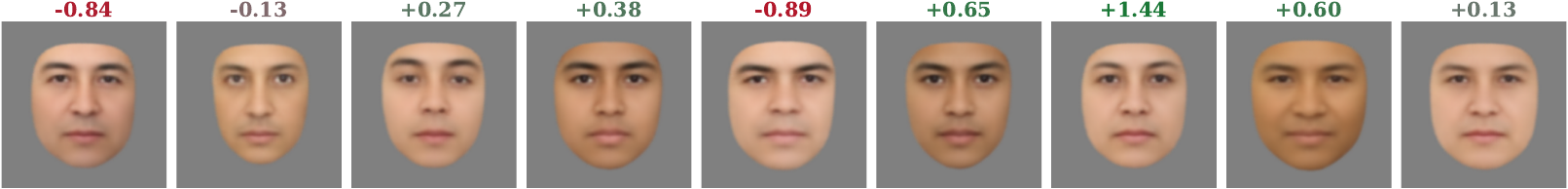}}%
  }
\end{figure}

\section{Methods}
\label{sec:methods}

To test the proposed variational interpretation of processing fluency, we train convolutional VAEs on face datasets without attractiveness supervision and evaluate whether human attractiveness judgments align with the resulting rate-distortion geometry and latent representations. The experimental design is summarized below, with an extended description of the methodology provided in Appendix \ref{app:methods}.

\subsection{Datasets}
\label{sec:data}

We trained unsupervised face representations from four large face image datasets: FairFace \citep{karkkainen2021fairface}, FFHQ \citep{karras2019style}, CelebA \citep{liu2015deep}, and UTKFace \citep{zhang2017age}. These datasets differ substantially in demographic composition, image quality, and pose variability, and therefore allow us to test whether the learned representational geometry generalizes across training distributions.

Human attractiveness judgments were evaluated on the Chicago Face Database (CFD) \citep{ma2015chicago}. We used the $597$ neutral-expression face images with available attractiveness ratings ($A$). CFD contains systematic differences in both images and ratings across demographics. All analyses were conditioned on eight separate demographic groups: four ethnicities (Asian, Black, Latino, White) and two genders (Female, Male). Therefore, all values were standardized to within-group zero-mean and unit variance z-scores.

\subsection{Preprocessing}

All training and evaluation images were processed by the same pipeline. Facial landmarks were detected using MediaPipe Face Landmarker \citep{lugaresi2019mediapipe}, and each image was transformed with a translation, rotation, and uniform scaling, such that the midpoint between the eyes and the center of the mouth lay at fixed coordinates within the $224\times224$ output image. The relative morphology of the face was fully preserved. Pixels outside the face region detected by MediaPipe were replaced by a gray color (see Fig.~\ref{fig:examples}). Images for which the required facial landmarks could not be reliably detected, e.g. because of occlusions or large rotations, were discarded.

\subsection{Variational face representations}
\label{sec:vae}

We trained convolutional VAEs independently on each of the four training datasets. The encoder consisted of five convolutional layers that compressed the $3\times224\times224$ RGB image to a $512\times7\times7$ representation, followed by two separate linear projections for the posterior mean $\mu_\phi$ and log variance $\log\sigma_\phi^2$, combined via the reparametrization trick \citep{kingma2013auto} into a linear latent representation space with a standard Gaussian prior and a baseline dimension size $d_z=128$. For each training dataset, we trained three independent models from different random initializations, giving $12$ baseline representations in total. Baseline models used $\beta=1$ and a fixed Gaussian decoder scale $\sigma=1$, and were trained for $50$ epochs with a learning rate of $2\times10^{-4}$, and KL warm-up over the first $10$ epochs. 

After training, all model parameters were frozen and the rates, distortions, and latent representations of the unseen 597 CFD faces were recorded for each model. The distortion for each face and model was computed from a Monte Carlo estimate with 16 samples.

\begin{figure}[!t]
\floatconts
  {fig:subfigex}
  {\caption{ \label{fig:alignment}
  Alignment between human attractiveness judgments and processing fluency in VAEs. \textbf{(a)} Relationship between human ratings and VAE ELBO for representations learned from CelebA. Each point denotes one CFD identity and the green line shows the linear regression with its 95\% bootstrapped confidence interval. \textbf{(b)} Distribution of human attractiveness in the rate-distortion space. Each point denotes one CFD identity colored by its human attractiveness rating. The background shows a descriptive attractiveness landscape, with red indicating lower and green indicating higher attractiveness. \textbf{(c)} Alignment between the empirical direction of increasing human attractiveness (gray) and the direction of increasing ELBO (green) for VAEs trained on different datasets.
  }}
  {%
    \subfigure[]{%
      \includegraphics[width=0.29\linewidth]{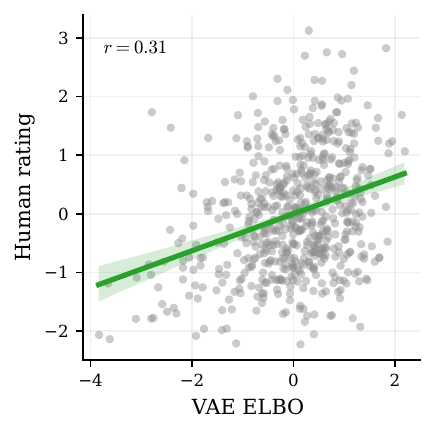}}%
    \hfill
    \subfigure[]{\label{fig:image-b}%
      \includegraphics[width=0.35\linewidth]{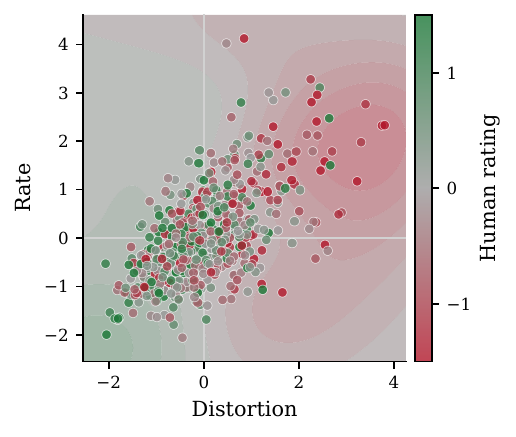}}
      \hfill
    \subfigure[]{\label{fig:image-b}%
      \includegraphics[width=0.29\linewidth]{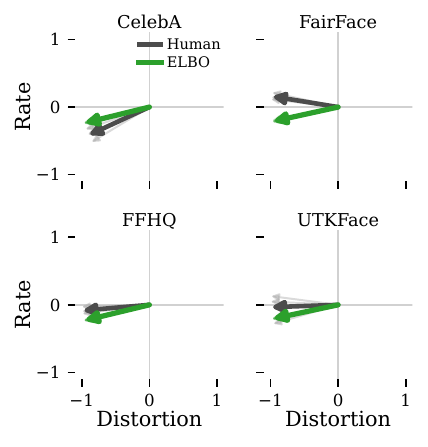}}
  }
\end{figure}

\section{Results}

\subsection{Attractiveness aligns with VAE's ELBO in rate-distortion space} \label{sec:results_elbo}

We first inquired whether human attractiveness judgments are systematically organized in the rate-distortion space of the learned variational face representations. Across all four training distributions, attractiveness was positively correlated with the ELBO (Fig.~\ref{fig:alignment}) with Pearson correlations of $r=0.312\pm.006$ for CelebA, $0.249\pm.006$ for FairFace, $0.285\pm.006$ for FFHQ, and $0.273\pm.009$ for UTKFace (all $p<0.001$). The association was significant for every individual baseline model ($r=0.244$ to $0.317$, $p<0.001$), despite the models having been trained independently and without aesthetic supervision. The two terms of the ELBO showed complementary associations with attractiveness. Lower distortion was consistently associated with higher attractiveness, with correlations $r= -0.257$ to $-0.300$ ($p<0.001$). Lower rate was also associated with higher attractiveness in every baseline model: $r=-0.117$ to $-0.256$ ($p<0.01$). Thus, attractive faces were both reconstructed more accurately and represented at lower rate. Across all $8$ ethnicity$\times$gender groups and $12$ baseline models, the within-group correlation between attractiveness and ELBO was positive in all $96/96$ combinations. The distortion correlations were likewise positive in $96/96$ combinations, while the rate associations were positive in $94/96$. The within-group correlations between attractiveness and ELBO, averaged across all models, ranged from $r=0.171$ (Latino-Female) to $0.431$ (Latino-Male). 

We next tested the geometric prediction that the direction of increasing attractiveness in rate-distortion space should align with the direction defined by the variational objective itself. Let $\bar{D}(x)$ and $\bar{R}(x)$ denote the within-group standardized rate-distortion coordinates of a face $x$. For each model, we estimated the empirical attractiveness direction by optimizing the parameters $a_D$, $a_R$, and $\epsilon$ to fit the relation
\begin{equation}
    \tilde{A}{(x)}
    =
    a_D D(x)
    +
    a_R R(x)
    +
    \epsilon ,
\end{equation}
resulting in a direction vector $\mathbf{a}=(a_D,a_R)$. Likewise, the ELBO direction in these coordinates is given by $\mathbf{e}\propto(-s_D,-s_R)$, where $e_D$ and $e_R$ are the residual scales of distortion and rate before standardization (see Appendix~\ref{app:rate_distortion_space}). We quantified alignment directly by the cosine similarity
\begin{equation}
    \cos\theta
    =
    \frac{
        \mathbf{a}^\top \mathbf{e}
    }{
        \|\mathbf{a}\|
        \|\mathbf{e}\|
    } ,
\end{equation}
where a cosine similarity of 1 indicates perfect agreement between the directions of increasing attractiveness and increasing ELBO. Our results showed a consistent agreement across all trained models. The per-dataset cosine similarities were $0.977\pm.023$ for CelebA, $0.929\pm.021$ for FairFace, $0.988\pm.008$ for FFHQ, and $0.969\pm.027$ for UTKFace. Across all $12$ individual models, $\cos\theta$ ranged from $0.905$ to $0.997$, corresponding to angular deviations of only $4.4^\circ$ to $25.2^\circ$. Therefore, the empirical attractiveness gradient consistently pointed toward the region of simultaneously lower distortion and lower rate selected by the variational objective.

Finally, we examined whether the result depended on the baseline parameter choice. Using FairFace as a controlled training distribution, we varied latent dimensionality ($d_z\in\{64,128,256,512\}$), regularization bottleneck ($\beta\in\{0.25,0.5,1,2,4\}$), and Gaussian decoder scale ($\sigma\in\{0.5,0.707,1,1.414\}$), with three independent seeds per configuration. Across all these different representation models, the mean correlation between attractiveness ELBO remained positive ($0.199$ to $.254$, $p<0.001$), while the mean cosine similarity between the empirical attractiveness gradient and the ELBO direction remained high ($0.896$ to $0.998$). The correspondence between human attractiveness and variational geometry was therefore not specific to any particular dataset or configuration but rather a property of the variational representation learning. 

\begin{figure}[!t] 
\floatconts
  {fig:subfigex}
  {\caption{ \label{fig:latent}
  Attractiveness geometry is conserved across independently learned latent spaces beyond prototypicality. \textbf{(a)} Two-dimensional projection of the latent spaces onto the attractiveness direction (horizontal) and the first principal component of the remaining orthogonal directions. \textbf{(b)} Pairwise cosine similarity (left) and cross-model transfer (right) between pairs of attractiveness directions in the 12 aligned VAE latent spaces. \textbf{(c)} Held-out predictions from different attractiveness sources.
  }} 
  {%
  \subfigure[Visualization of the attractiveness-aligned latent geometries.]{%
    \includegraphics[width=0.99\linewidth]{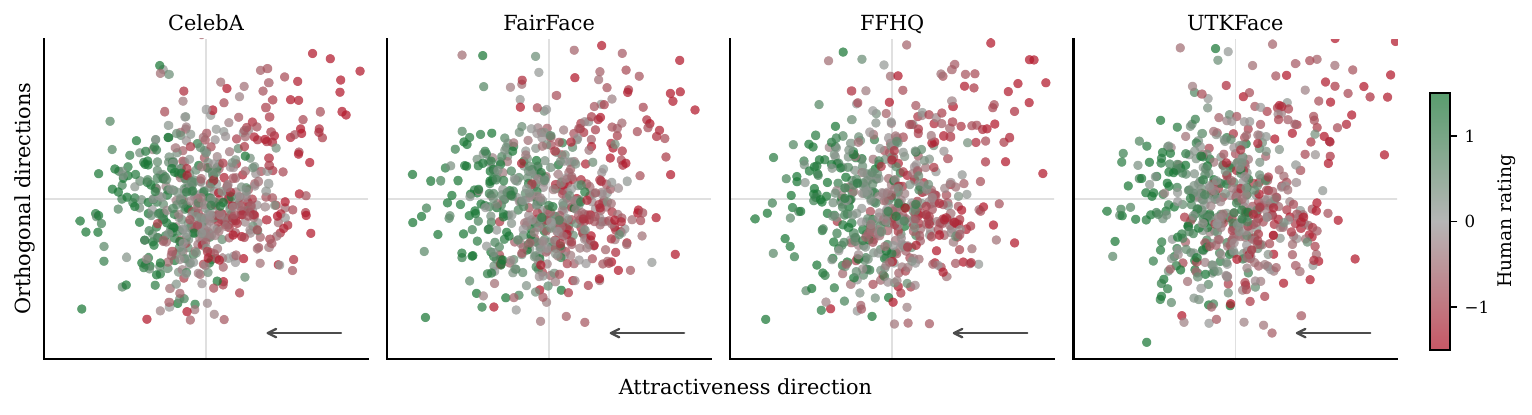}} 

    \vspace{1em}
         
    \subfigure[Correspondence in pairs of models]{%
      \includegraphics[width=0.66\linewidth]{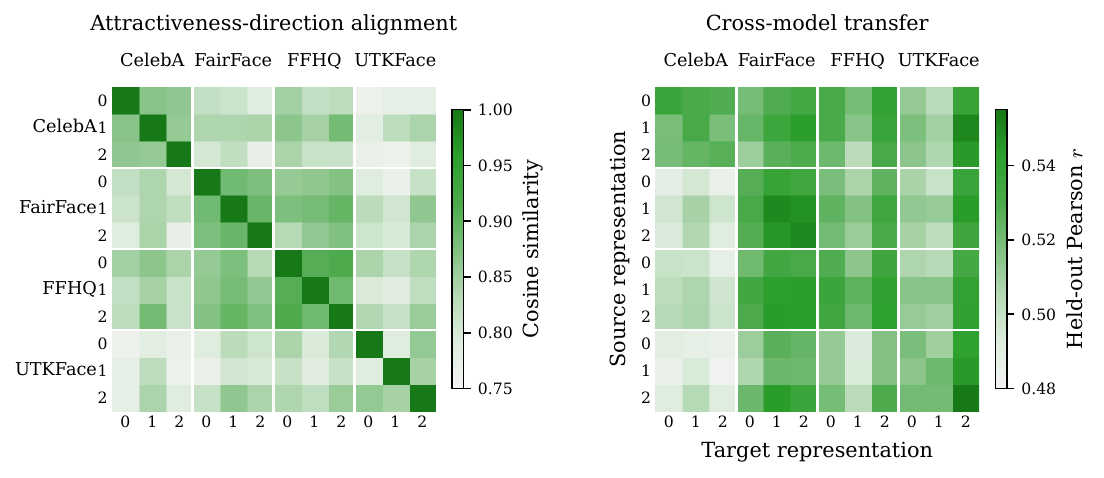}}%
    \hfill
    \subfigure[Prediction sources]{\label{fig:image-b}%
      \includegraphics[width=0.31\linewidth]{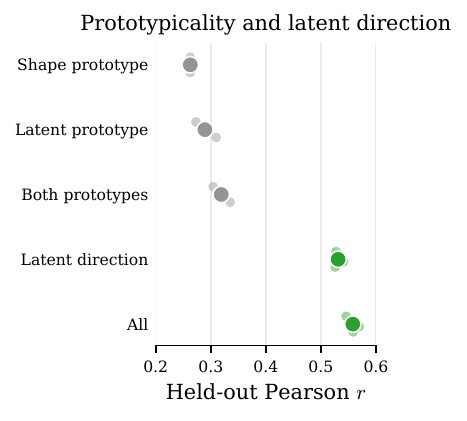}}
  }
\end{figure}

\subsection{Attractiveness geometry is conserved across different latent spaces}
\label{sec:results_latent_geometry}

The previous analysis established a common relationship between attractiveness and the rate-distortion space of the VAEs. We next asked whether these independently trained VAEs also organize their face codes along a shared latent attractiveness direction. We compared the mean posterior representations $\mu_\phi(x)\in\mathbb{R}^{128}$ relative to their demographic group centroids by aligning the $12$ latent spaces via generalized orthogonal Procrustes. For each model $m$, we estimated a linear direction $\mathbf{A}_m$ associated with increasing attractiveness. We then measured the cosine similarity between these directions across every pair of independently trained models. The resulting directions were strongly conserved (Fig.~\ref{fig:latent}). Across all $66$ model pairs, the mean pairwise cosine similarity was $0.833\pm0.030$, and every pair was positively aligned, with similarities ranging from $0.764$ to $0.914$. Directions learned from different random seeds of the same training distribution were especially similar ($0.870\pm0.032$), and the effect remained strong when comparing models trained on different face datasets ($0.825\pm0.034$). For comparison, we permuted the attractiveness labels within demographic groups while preserving the correspondence of faces across representations and found that the mean pairwise direction cosine was $0.632\pm0.038$, reflecting the shared structure of the aligned face manifolds but well below the attractiveness alignment. 

Next, to further investigate the correspondence between attractiveness directions among pairs of models, we performed a cross-model transfer analysis with five repeats of a five-fold cross-validation. The learned direction from one model (source) was transported into another (target) latent space and used to predict attractiveness scores on the held-out faces. At no point were the attractiveness ratings or latent representations of held-out identities used to estimate the alignment or direction. When the source and target were the same model, i.e. no cross-model transfer, the baseline correlation with attractiveness was $r=0.534\pm.012$ across the $12$ models. Replacing the source representation by an independently initialized model trained on the same dataset produced almost no degradation ($r=0.529\pm.011$). Most importantly, transfer remained nearly as strong when source and target models were trained on different face distributions with mean held-out correlation $r=0.516\pm.017$. In other words, the human attractiveness dimension identified in one learned representation can be transported to another representation with little loss of predictive information. This suggests that human attractiveness judgments correspond to a reproducible direction embedded in the higher-dimensional geometry learned by generative models of faces.

\subsection{Attractive faces have prototypical shapes and latent codes}
\label{sec:results_prototypicality}

We next inquired whether attractiveness could be explained by a simpler account based on the prototypicality of the faces. Classic theories of face perception have emphasized that faces closer to a population prototype are often judged as more attractive. We first quantified classical shape prototypicality from the $468$ facial landmarks detected by MediaPipe. Each face was compared with a leave-one-out prototype computed from the remaining members of the same demographic group. Shape prototypicality was defined as the negative Procrustes distance to this within-group prototype, such that larger values indicate more prototypical facial shape. We found that more prototypical faces were rated as more attractive ($r=0.268, p<0.001)$. This replicated the classical averageness effect \citep{winkielman2006prototypes,ryali2020likely} within the CFD faces.  

We then asked whether the same relationship emerges in the latent representations learned by the VAEs. For each model $m$, latent prototypicality was defined as the negative Euclidean distance between the mean posterior representation $\mu_{\phi_m}(x)$ and the leave-one-out centroid of the corresponding demographic group. Attractiveness was positively associated with latent prototypicality in every model, with Pearson correlations of $r=0.315\pm.002$ for CelebA, $0.294\pm.013$ for FairFace, $0.294\pm.004$ for FFHQ, and $0.281\pm.012$ for UTKFace. Thus, the tendency for attractive faces to lie closer to the within-group prototypes is reproduced not only in explicit landmark geometry but also in independently learned generative representations. The two notions of prototypicality were related but not redundant. Across models, the correlation between landmark-space and latent-space prototypicality ranged from $r=0.455$ to $0.544$. Hence, the learned representations preserve a substantial component of classical facial averageness, while also reorganizing the faces according to additional image structure not captured by the landmark prototype alone (e.g., skin tone or eye color).

Finally, we found that the projections onto the latent attractiveness directions identified in Sec.~\ref{sec:results_latent_geometry} were substantially more predictive of attractiveness than either prototype measure (see Fig.~\ref{fig:latent}). Across training datasets, the latent direction alone predicted held-out attractiveness with correlations of $r=0.526$ to $0.541$. The full model containing shape prototypicality, latent prototypicality, and the latent attractiveness direction reached only moderately higher correlations ($r=0.546$ to $0.570$). In sum, most of the predictable attractiveness structure captured by these representations lies along the learned latent direction rather than being reducible to either notion of prototypicality.

\section{Discussion}
\label{sec:discussion}

This work investigated whether human judgments of facial attractiveness are reflected in the structure of independently learned variational representations of faces. Across four training datasets, our results converge on a set of common findings. First, attractiveness increases along the direction of higher processing fluency: attractive faces occupy regions characterized by lower reconstruction distortion and lower regularization cost. Second, the relationship extends beyond the scalar ELBO objective. After aligning independently learned latent spaces, attractiveness defines a direction that transfers across models and training datasets with little loss of predictive information. Third, attractive faces are more prototypical in both landmark and latent space, but this prototype effect accounts for only part of the latent attractiveness geometry. Together, these findings suggest that classical findings of facial attractiveness can be understood as manifestations of a representational organization resulting from unsupervised learning of a generic generative model.

The two components of the ELBO, rate and distortion, could hint at two distinct sources for processing fluency. The distortion is a measure of the fidelity with which a stimulus is represented whereas the rate quantifies the complexity of that representation. We found that both components correlate with attractiveness, although it is their trade-off in the ELBO that is best aligned with attractiveness ratings. Our variational account can thus integrate two complementary interpretations  of processing fluency, identified by \citet{brielmann2022computational} as the \textit{a priori} and \textit{a posteriori} signals. In this view, a low rate indicates that the representation remains close to the model's expectations and a low distortion reflects that the stimulus was correctly processed. However, this correspondence remains preliminary and will require empirical validation to disentangle the two sources of processing fluency.  

While our results support the hypothesis that faces are represented with ease when they can be reconstructed accurately while deviating minimally from the latent prior, this interpretation should not be taken to imply that human observers explicitly optimize an ELBO. Instead, the ELBO provides a computationally explicit measure of processing fluency in a learned generative representation. The most surprising result is that a direction determined entirely by the model's unsupervised objective is closely aligned with human preference. In Appendix \ref{app:results} we report that objective facial properties (such as the distance between the eyes) predicted both human attractiveness and VAE ELBO with remarkable similarity ($r=0.509$ and $0.499$, respectively), whereas social and affective judgments were much more predictive of human attractiveness ($r=0.704$) than of VAE ELBO ($r=0.262$). Therefore, human aesthetic judgments can only partially be captured by an unsupervised image model trained passively on face images.

Several limitations constrain the conclusions of the present study. The VAEs considered here are deliberately simple generative models and should not be interpreted as mechanistic models of the human visual system. The CFD dataset contains frontal neutral-expression images, which makes it well suited to isolate facial structure but limits its generalization to naturalistic viewing conditions, pose variability, and dynamic face expressions. Our demographic conditioning reduces obvious within-group confounding effects but relies on coarse dataset categories and does not capture the full diversity of individual experience or cultural preference. The attractiveness ratings are also population averages and do not capture the variability of individual differences. Finally, it must be noted that all reported relationships are correlational and not causal. Our results can only establish a correspondence between human preference and learned representational geometry. Nevertheless, this correspondence provides empirical computational support for the a variational interpretation of processing fluency in face perception, linking human preference to the structure of generative representations.

\acks{
 Experiments were performed on the computational cluster Katana supported by Research Technology Services at UNSW Sydney (DOI: 10.26190/669X-A286). This work was supported by the Deutsche Forschungsgemeinschaft (German Research Foundation, DFG) under Germany's Excellence Strategy (EXC 3066/1 ``The Adaptive Mind'', Project No. 533717223). J.T. was supported by the Johanna Quandt foundation.
}

\bibliography{1_REFERENCES}

\newpage
\appendix

\section{Extended methods}\label{app:methods}

\subsection{Dataset details}

\subsubsection{Training data}

We trained face representations from four datasets with different demographic distributions and image statistics. Importantly, all available labels and metadata were ignored during training. The models learned face representations fully unsupervised.

\paragraph{CelebA} \citep{liu2015deep} contains 202,599 face images (163,456 retained after preprocessing) of  celebrity identities collected from the internet. This was the largest dataset used in our experiments, and its interest is that it provides substantial variation in pose, appearance, and background. However, CelebA was not constructed to be demographically balanced, and must thus be seen as a large in-the-wild distribution of faces with implicit biases. 

\paragraph{FairFace} \citep{karkkainen2021fairface} contains 108,501 face images (52,801 retained after preprocessing), collected primarily from Flickr and subsequently annotated for race, gender, and age. It was explicitly constructed to reduce the strong racial imbalance present in CelebA and other popular face datasets. FairFace distinguishes seven race categories (Black, East Asian, Indian, Latino, Middle Eastern, Southeast Asian, and White), two gender categories, and nine age ranges. These annotations were not used during training.

\paragraph{FFHQ} \citep{karras2019style} contains 70,000 high-quality $1024\times1024$ face images (57,741 retained after preprocessing) collected from Flickr. The dataset was designed for generative face modeling and deliberately includes broad variation in age, ethnicity, and background. FFHQ provides high-quality images beyond the ranges of the other datasets, although this property is mostly lost since we resized the images to $224\times224$ pixels.

\paragraph{UTKFace} \citep{zhang2017age} contains more than 20,000 internet face images (18,492 after preprocessing) covering a large range of identities. It is particularly interesting because it covers ages from $0$ to $116$ years with five ethnicity categories (Asian, Black, Indian, White, and Other). In line with the other training datasets, the annotations were ignored during training.

\subsubsection{Evaluation data}

\paragraph{Chicago Face Database (CFD)} \citep{ma2015chicago} contains standardized face images from 597 individuals. The participants self-identified as Asian, Black, Latino, or White and as Female or Male, yielding eight ethnicity$\times$gender groups. The dataset contains multiple expressions but we used exclusively the neutral-expression images. All 597 images were retained after preprocessing. CFD provides both objective measurements of facial morphology and subjective ratings, including attractiveness, collected from an average of 25 participants per image on a scale between 1 and 7.

No CFD images or ratings were used during the training of the VAEs or for hyperparameter selection. Facial statistics and attractiveness ratings were found to differ between demographics, particularly with Female faces yielding significantly higher scores. To standardize our analysis, every scalar quantity $y_i$ (e.g. the attractiveness rating) was transformed as
\begin{equation}
    y_i^{(z)}
    =
    \frac{
        y_i-\mu_{g(i)}
    }{
        \sigma_{g(i)}
    },
\end{equation}
where $\mu_{g(i)}$ and $\sigma_{g(i)}$ are the mean and standard deviation among CFD faces belonging to the same demographic group $g(i)$ as face $i$.

\subsection{Face preprocessing}
\label{app:preprocessing}

\begin{figure}[!t]
\floatconts
  {fig:subfigex}
  {\caption{ \label{fig:preprocessing}
  Example of face preprocessing pipeline with an image from the FFHQ dataset. \textbf{(a)} Original image. \textbf{(b)} Facial geometry detected by MediaPipe Face Landmarker. \textbf{(c)} Image after translation, rotation, and uniform scaling. \textbf{(d)} Final $224\times224$ input after masking the non-face region.
  }}
  {%
    \subfigure[]{%
      \includegraphics[width=0.24\linewidth]{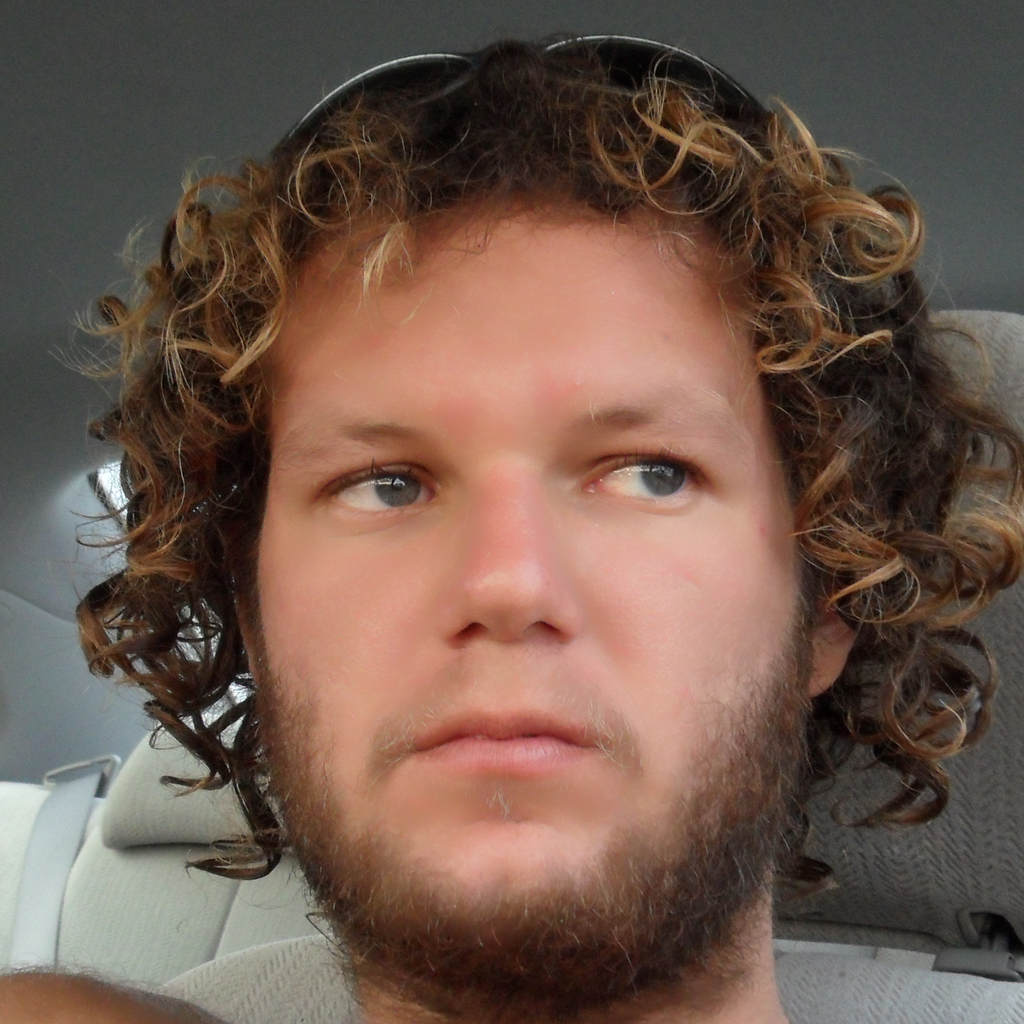}}%
    \hfill
    \subfigure[]{\label{fig:image-b}%
      \includegraphics[width=0.24\linewidth]{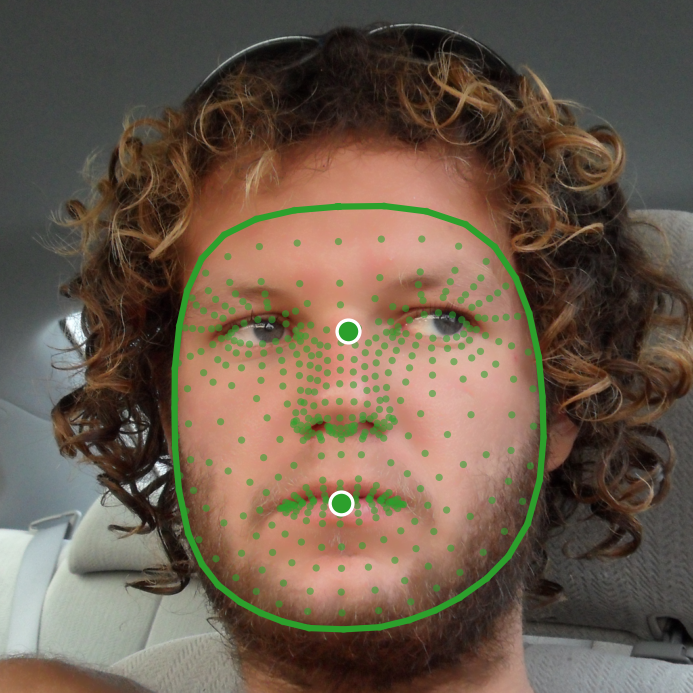}}
      \hfill
    \subfigure[]{\label{fig:image-b}%
      \includegraphics[width=0.24\linewidth]{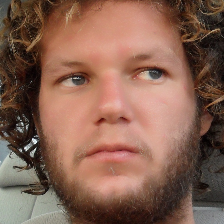}}
        \hfill
    \subfigure[]{\label{fig:image-b}%
      \includegraphics[width=0.24\linewidth]{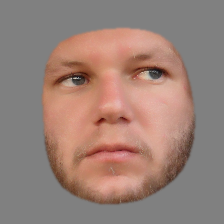}}
  }
\end{figure}

All five image datasets were preprocessed with the same pipeline before training or evaluation. First, we detected facial geometry using MediaPipe Face Landmarker \citep{lugaresi2019mediapipe}, which provides a dense set of $468$ facial landmarks. Images for which the eyes, mouth, or complete face outline could not be localized were rejected. Additionally, images were the face was not facing forward were also rejected.  For each accepted image, we applied a transformation containing translation, rotation, and uniform scaling, such that the midpoint between the two eyes was located at $(0.50W,0.35H)$ and the mouth center at $(0.50W,0.68H)$ in the final coordinate frame. No warping or anisotropic scaling was used. This procedure standardized global face position and size while preserving individual facial morphology, e.g. the relative separation of the eyes with respect to the distance between the eyes and the mouth was maintained.  After the alignment, the facial region was detected with the MediaPipe face-oval contour, and all pixels outside the resulting facial mask were replaced by the constant RGB value $(128,128,128)$. The final images were stored as $224\times224$ RGB arrays. Fig.~\ref{fig:preprocessing} illustrates the complete procedure.

\subsection{VAE architecture}
\label{app:vae_architecture}

All experiments used the same convolutional VAE architecture (see Table~\ref{tab:vae_architecture}). The model received a $224\times224$ RGB image $x\in[0,1]^{3\times224\times224}$ and encoded it through five convolutional layers. The final $512\times7\times7$ tensor was flattened to a $25{,}088$-dimensional vector and projected independently onto the mean $\mu_\phi(x)$ and log-variance $\log \sigma^2_\phi(x)$ of a diagonal Gaussian approximate posterior. The baseline latent dimensionality was $d_z=128$ and the prior was a standard Gaussian $p(z)=\mathcal{N}(0,I)$. The decoder mirrored the encoder. The latent samples were obtained with the reparameterization trick \citep{kingma2013auto}:
\begin{equation}
    z =
    {\mu}_\phi(x)
    +
    {\sigma}_\phi(x)
    \odot
    \epsilon,
    \qquad
    \epsilon\sim\mathcal{N}(0,I).
\end{equation}

\begin{table}[ht]
\centering
\caption{
Architecture of the convolutional VAE used in our experiments. All convolutions and transposed convolutions used kernel size $4$, stride $2$, and padding $1$. The baseline latent dimensionality was $d_z=128$.
}
\label{tab:vae_architecture}
\small
\begin{tabular}{llll}
\toprule
Stage & Layer & Output shape & Activation \\
\midrule
Input
    & -- 
    & $3\times224\times224$
    & -- \\

\midrule
Encoder 1
    & Conv2d $3\rightarrow32$
    & $32\times112\times112$
    & ReLU \\

Encoder 2
    & Conv2d $32\rightarrow64$
    & $64\times56\times56$
    & ReLU \\

Encoder 3
    & Conv2d $64\rightarrow128$
    & $128\times28\times28$
    & ReLU \\

Encoder 4
    & Conv2d $128\rightarrow256$
    & $256\times14\times14$
    & ReLU \\

Encoder 5
    & Conv2d $256\rightarrow512$
    & $512\times7\times7$
    & ReLU \\

Flatten
    & --
    & $25{,}088$
    & -- \\

Posterior mean
    & Linear $25{,}088\rightarrow128$
    & $128$
    & -- \\

Posterior log-var.
    & Linear $25{,}088\rightarrow128$
    & $128$
    & -- \\

\midrule
Latent sample
    & Reparameterization
    & $128$
    & -- \\

\midrule
Projection
    & Linear $128\rightarrow25{,}088$
    & $512\times7\times7$
    & -- \\

Decoder 1
    & ConvTranspose2d $512\rightarrow256$
    & $256\times14\times14$
    & ReLU \\

Decoder 2
    & ConvTranspose2d $256\rightarrow128$
    & $128\times28\times28$
    & ReLU \\

Decoder 3
    & ConvTranspose2d $128\rightarrow64$
    & $64\times56\times56$
    & ReLU \\

Decoder 4
    & ConvTranspose2d $64\rightarrow32$
    & $32\times112\times112$
    & ReLU \\

Decoder 5
    & ConvTranspose2d $32\rightarrow3$
    & $3\times224\times224$
    & Sigmoid \\

\bottomrule
\end{tabular}
\end{table}

Unless otherwise specified, all experiments used $d_z=128$, $\beta=1$, and $\sigma_x=1$. For each of the four datasets, three models were trained from independent random initializations, yielding the 12 baseline representations analyzed throughout the main experiments. Additional experiments with the FairFace dataset retained the same architecture while varying latent dimensionality, regularization trade-off $\beta$, or decoder scale $\sigma$, as described in Sec.~\ref{sec:results_elbo}.

\subsection{Rate-distortion evaluation}

For each trained model and each CFD face $x_i$, the encoder produced a posterior distribution $q_\phi(z\mid x_i)=\mathcal{N}(\mu_i,\mathrm{diag}(\sigma_i^2))$. The rate was computed  analytically as
\begin{equation}
R_i
=
\frac{1}{2}
\sum_{j=1}^{d_z}
\left(
\mu_{ij}^2
+
\sigma_{ij}^2
-
\log\sigma_{ij}^2
-
1
\right).
\end{equation}

In parallel, the distortion was approximated by the Monte Carlo estimate 
estimated as
\begin{equation}
D_i
=
\mathbb{E}_{q_\phi(z\mid x_i)}
\left[
\frac{1}{2\sigma_x^2}
\|x_i-f_\theta(z)\|_2^2
\right],
\end{equation}
using $16$ independent posterior samples.

\subsection{ELBO in rate-distortion space}
\label{app:rate_distortion_space}

Our first experiment (Sec.~\ref{sec:results_elbo}) tested whether attractiveness is organized along the rate-distortion trade-off induced by Eq.~\eqref{eq:elbo}. For every model, we computed the distortion $D_i$ and the rate $R_i$ for all CFD faces. To test the prediction that attractiveness follows the direction selected by the variational objective, we standardized the within-group residual distortion and rate coordinates,
\begin{equation}
    \bar{D}_i
    =
    \frac{D_i-\mathbb{E}[D_i\mid g_i]}{s_D},
    \qquad
    \bar{R}_i
    =
    \frac{R_i-\mathbb{E}[R_i\mid g_i]}{s_R},
\end{equation}
where $s_D$ and $s_R$ denote the residual standard deviations, so that the direction of the ELBO in standardized rate-distortion coordinates is given by 
\begin{equation}
    \mathbf{e}
    \propto
    (-s_D,-s_R).
\end{equation}

We also estimated the empirical attractiveness gradient with the linear model
\begin{equation}
    \tilde{A}
    =
    a_D \bar{D}
    +
    a_R \bar{R}
    +
    \epsilon,
\end{equation}
from which we define the attractiveness direction
\begin{equation}
    \mathbf{a}_A=(a_D,a_R).
\end{equation}

Finally, the agreement between the empirical attractiveness direction and the ELBO was measured as the cosine similarity between both vectors.

\subsection{Alignment and transfer of latent attractiveness geometry}
\label{sec:methods_latent}

Our second analysis asked whether attractiveness is associated with a common direction in the latent representations learned by independent VAEs. For model $m$, we represented every CFD face by its posterior mean
\begin{equation}
    \mathbf{z}^{(m)}_i
    =
    \mu_{\phi_m}(x_i)
    \in\mathbb{R}^{128}.
\end{equation}
Within each ethnicity$\times$gender group we subtracted the group-specific latent centroid. First, we aligned the $12$ representations using generalized orthogonal Procrustes analysis. Specifically, for centered representation matrices $Z_m$, we estimated orthogonal transformations $Q_m$ with respect to a common template $M$ (also optimized) by minimizing a Procrustes objective
\begin{equation}
    \mathcal{L} = \sum_m
    \left\|
        Z_mQ_m-M
    \right\|^2,
    \qquad
    \text{subject to }Q_m^\top Q_m=I.
    \label{eq:procrustes_latent}
\end{equation}

After alignment, we estimated the linear attractiveness direction and measured the similarity between every pair of model directions via their cosine similarity. To compare with a ``default alignment baseline'', we constructed a within-group permutation where the attractiveness ratings were shuffled relative to the identities. We repeated the procedure with $1000$ permutations.

\subsection{Prototypicality and geometric representability}
\label{sec:methods_proto}

Our third analysis tested whether the attractiveness effects above could be explained from the likeness of each face to its within-group prototype. We considered prototypicality independently in landmark and latent space. First, each face was represented by the $468$ MediaPipe facial landmarks (see Fig.~\ref{fig:preprocessing}). The landmarks were aligned with a two-dimensional orthogonal Procrustes transformation. We subsequently computed the leave-one-out within-group prototype $\bar S_{g,-i}$ of identity $i$. Prototypicality was defined as the negative Procrustes distance to the prototype. Additionally, for each trained VAE, latent prototypicality was analogously defined from posterior means as the negative Euclidean distance in latent space to the leave-one-out centroid of the corresponding group. We also measured the relationship between shape and latent prototypicality as their linear correlation to determine how closely the learned notion of latent prototypicality corresponded to classical facial shape averageness.

We also quantified how faithfully each VAE preserved facial geometry. For each CFD face, we decoded the posterior mean, $\hat x_i = f_\theta (\mu_\phi(x_i))$, and extracted its facial landmarks using MediaPipe on the reconstruction $\hat x_i$. We then computed the Procrustes distance between the original landmarks and the reconstructed landmarks. Higher distances indicate less faithful preservation of facial structure.

Finally, we tested whether the latent attractiveness direction carries predictive information beyond prototypicality. We compared linear models based on shape prototypicality, latent prototypicality, latent attractiveness-direction, and their combinations. All prototypes, demographic centroids, normalization parameters, and attractiveness directions were estimated from training identities only. We computed the correlation between held-out predictions and actual attractiveness ratings. 

\subsection{Statistics}
\label{app:statistics}

Unless otherwise stated, values reported across models are means and standard deviations across the three independent seeds for each training distribution. All within-group demographic standardization used only the ethnicity$\times$gender labels supplied by CFD. For descriptive analyses (correlations, alignment), statistics were computed from the complete evaluation set. For the predictive cross-validation analyses all the values were estimated exclusively from training identities and subsequently applied without any changes to the held-out identities, on which the results and statistics were computed.

\section{Supplementary results} \label{app:results}

\begin{figure}[!t]
\floatconts
  {fig:robustness}
  {\caption{
    Robustness of attractiveness-ELBO alignment to different hyperparameters. (\textbf{A}) Pearson correlations attractiveness ratings and VAE ELBO. (\textbf{B}) Cosine similarity between attractiveness and ELBO directions in rate-distortion space.
  }}
  {%
    \includegraphics[width=\linewidth]{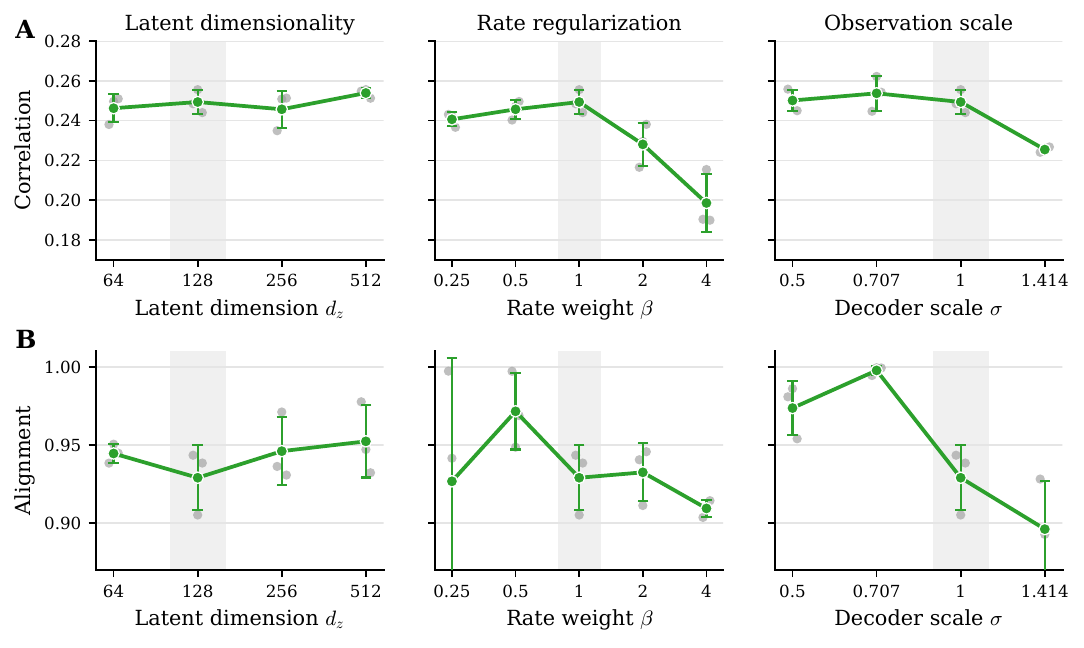}%
  }
\end{figure}

\subsection{Robustness across hyperparameters}

Using the FairFace dataset, we tested whether the correlation between attractiveness and VAE ELBO and the alignment of their directions in rate-distortion space was conserved across different sets of hyperparameters. We varied one of the default model parameters ($d_z=128$, $\beta=1$, $\sigma=1$) at a time while holding all other components fixed across the ranges $d_z\in\{64,128,256,512\}$, $\beta\in\{0.25,0.5,1,2,4\}$, and $\sigma\in\{0.5,0.707,1,1.414\}$, all with three random seeds. The results are shown in Fig.~\ref{fig:robustness}. In all individual models, the correlation between human ratings and ELBO was significant, yielding Pearson correlations $r=0.190 $ to $r=0.262$, all $p<0.001$. The lowest average correlation was observed for $\beta=4$ with $r=0.199\pm0.015$, possibly due to an over-regularization of the latent space where the face reconstructions are too degraded, although the correspondence between attractiveness and processing fluency was retained. The lowest alignment between the directions of the attractiveness and ELBO vectors in rate-distortion space was observed for $\sigma=\sqrt{2}$ whereas the highest value was $0.998\pm0.003$ for $\sigma=1/\sqrt{2}$, where the two vectors were virtually collinear. 

\begin{figure}[!t]
\floatconts
  {fig:demographics}
  {\caption{
    Robustness of correlations across demographic groups. Each column represents a seed within a training dataset (see Fig.~\ref{fig:latent}).
  }}
  {%
    \includegraphics[width=\linewidth]{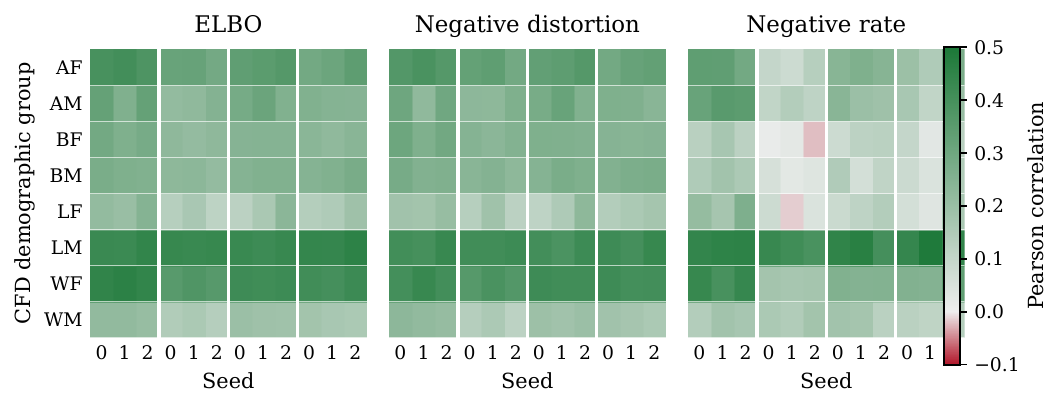}%
  }
\end{figure}

\subsection{Robustness across demographic groups}

To validate whether the relationship between attractiveness and processing fluency was consistent in all demographics (ethnicity$\times$gender), we independently computed the Pearson correlations between the human ratings and the ELBO, negative distortion, and negative rate within group for each of the 12 baseline models. In total, 96 comparisons per metric were computed. We found that the correlation between ELBO and attractiveness was positive in all 96 comparisons, with mean correlation $r=0.286\pm0.097$ and individual values ranging between $r=0.115$ and $0.457$. Likewise, the negative distortion was positively correlated with attractiveness across all 96 comparisons ($r=0.113$ to $0.431$). On the other hand, the negative rate was positively correlated in 94 of the 96 comparisons, with the lowest effect being virtually zero: $r=-0.022$. Interestingly, our results show considerable differences among different demographics. The highest within-group correlations between attractiveness and ELBO were achieved for the Latino-Male ($r=0.431$) and White-Female ($r=0.408$) groups, whereas the lowest correlations were achieved for the Latino-Female ($r=0.171$) and White-Male ($r=0.178$) groups. We speculate that these differences are more reflective of training and evaluation data imbalances than of objective properties of these particular demographics. Nevertheless, further experiments are required to validate this hypothesis.

\subsection{Preserved facial geometry}

We tested whether attractive facial geometry is represented more faithfully by the generative model, that is whether the VAEs preserve the shape of attractive faces more than unattractive faces in their reconstructions. For each reconstructed face, we extracted the same facial landmarks and measured the Procrustes discrepancy between the original and reconstructed shape. We found that landmark preservation was moderately associated with attractiveness in all $12$ models, with  correlations of $r=0.135\pm.005$ for CelebA, $0.138\pm.015$ for FairFace, $0.152\pm.007$ for FFHQ, and $0.158\pm.008$ for UTKFace. Attractive faces therefore exhibited greater latent prototypicality and more faithful reconstruction of the facial geometry, although the latter association was moderate.

\begin{figure}[!t]
\floatconts
  {fig:properties}
  {\caption{
    Correlations of human attractiveness judgments (gray) and VAE ELBO scores (green) against objective and subjective face properties.
  }}
  {%
    \includegraphics[width=\linewidth]{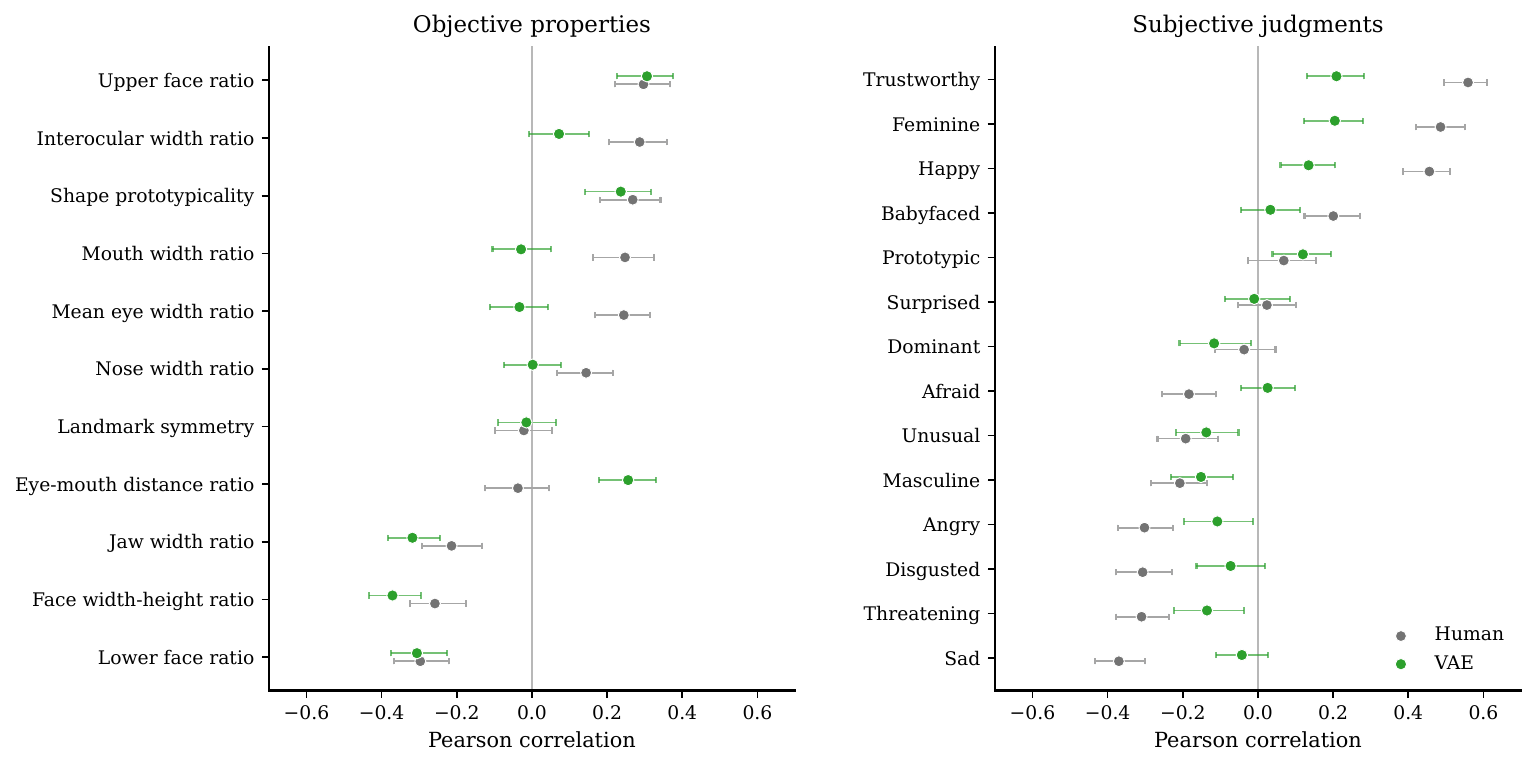}%
  }
\end{figure}

\subsection{Objective and subjective facial properties}

Finally, we investigated how different facial properties correspond to human attractiveness judgments and the VAE's processing fluency. The face properties were split into two separate categories. On the one hand, we considered objective geometric measurements derived from facial landmarks, such as the distance between the eyes (interocular width) or the aforementioned shape prototypicality. On the other hand, we evaluated other social or affective ratings included in the CFD: trustworthiness, happiness, femininity, disgust, etc. These judgments were reported by the same participants that judged attractiveness. Consequently, these subjective judgments were expected to be tightly associated.

The objective facial properties showed substantial agreement between human and VAE ratings. The correlations between the two correlation profiles shown in Fig.~\ref{fig:properties} was $r=0.760$, with some particular properties being closely matched. For example, the upper-face ratio correlated positively with both scores ($r=0.297$ for human and $r=0.306$ for ELBO). However, other geometric features deviated considerably, showing that attractiveness cannot be reduced to objective properties alone. As for the subjective judgments, the disagreement between data and model was larger. Faces were considered more attractive when they were more trustworthy, feminine, and happy, and less attractive when they were more threatening and sad. Interestingly enough, the VAE is able to show similar trends, albeit in a moderate effect. For example, trustworthiness had a Pearson correlation of $r=0.559$ with attractiveness but only $r=0.209$ with ELBO.

To further compare objective and subjective properties, we performed a ridge-regressions for each separate and for their combination and evaluated them with out-of-fold predictions. Objective properties predicted attractiveness and ELBO with comparable accuracy ($r=0.509$ and $0.499$, respectively). However, subjective properties were considerably better for human attractiveness ($r=0.704$) than for ELBO ($r=0.262$). The combination of all properties increased accuracies to $r=0.765$ for human ratings and $0.523$ for ELBO.

In sum, these results support our variational account of processing fluency insofar as the VAEs can capture the statistical regularities of the faces, whereas social and affective judgments remain unavailable to models trained with unsupervised learning. However, the moderate similarities between human ratings and VAE ELBO for subjective judgments points to a possible extension of this work beyond attractiveness and into other evaluations.

\newpage 
\section{Examples of face reconstructions}\label{app:reconstructions}

\begin{figure}[!h]
\floatconts
  {fig:reconstructions}
  {\caption{
    Reconstructions of Asian  Female (top) and Male (bottom) faces. Rows: (1) original CFD images, (2-5) Reconstructions from VAE trained on (2) FairFace, (3) FFHQ, (4) CelebA, (5) UTKFace. See Fig.~\ref{fig:examples} for details.  
  }}
  {%
    \includegraphics[width=.9\linewidth]{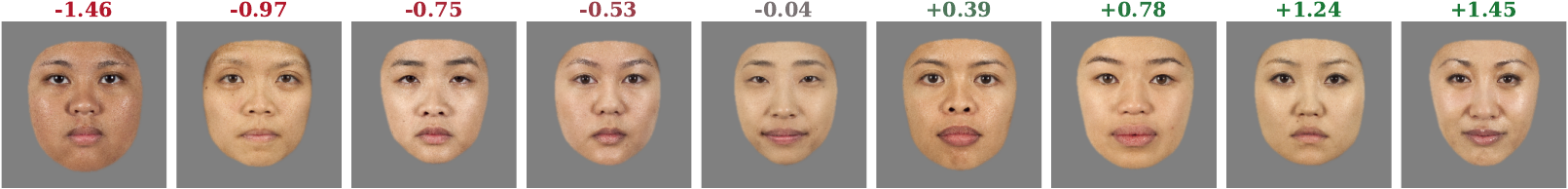}%
    
    \includegraphics[width=.9\linewidth]{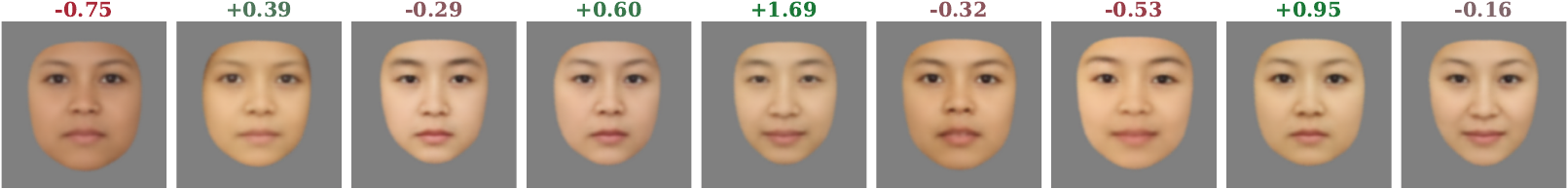}%
    
    \includegraphics[width=.9\linewidth]{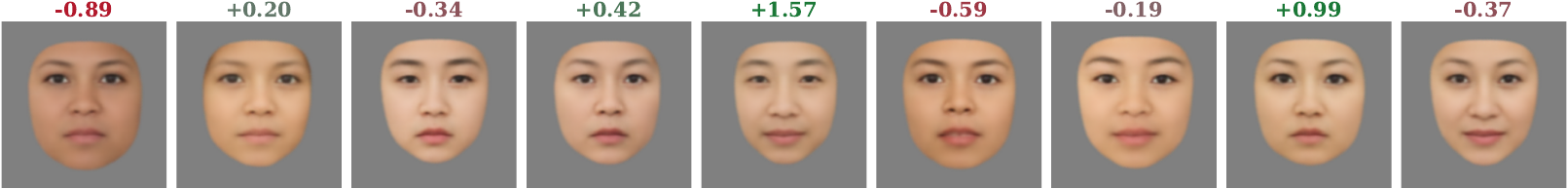}%
    
    \includegraphics[width=.9\linewidth]{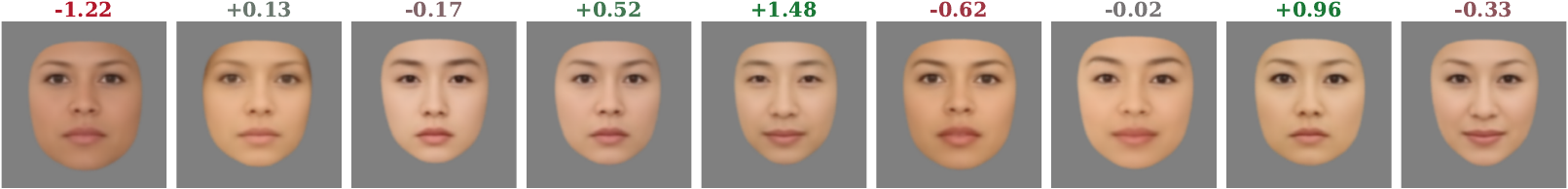}%
    
    \includegraphics[width=.9\linewidth]{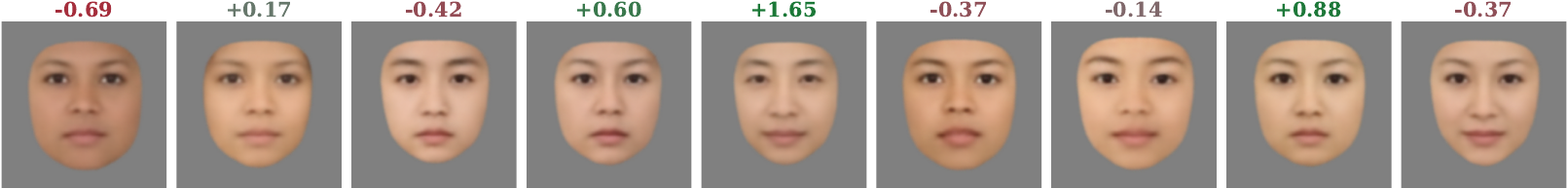}%

    \vspace{2em}

    \includegraphics[width=.9\linewidth]{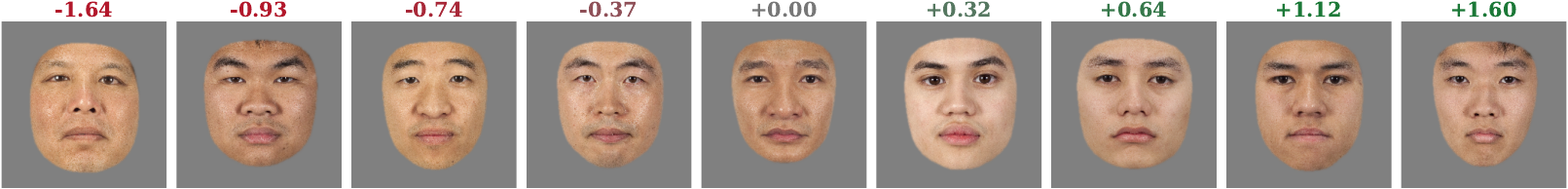}%
    
    \includegraphics[width=.9\linewidth]{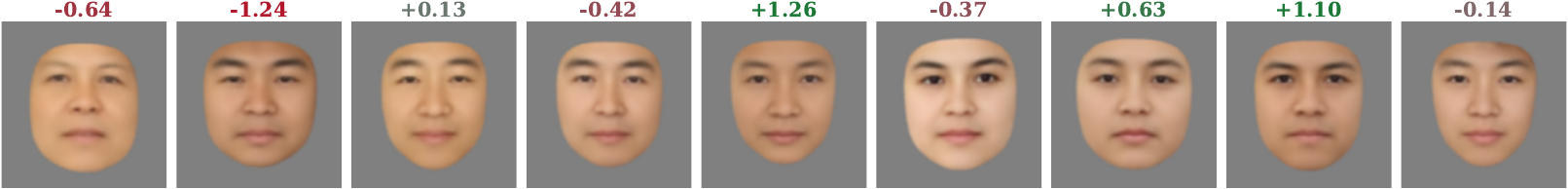}%
    
    \includegraphics[width=.9\linewidth]{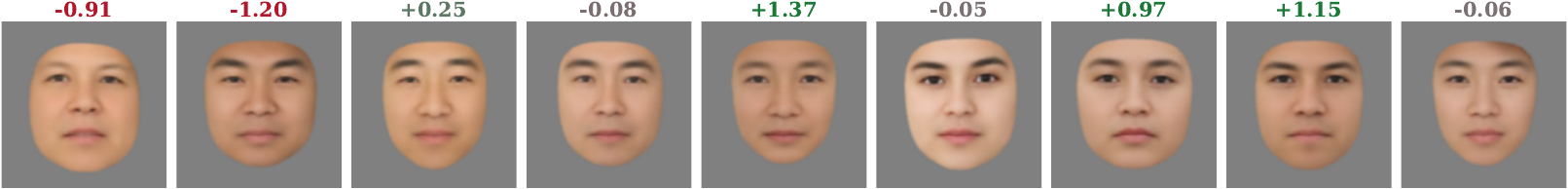}%
    
    \includegraphics[width=.9\linewidth]{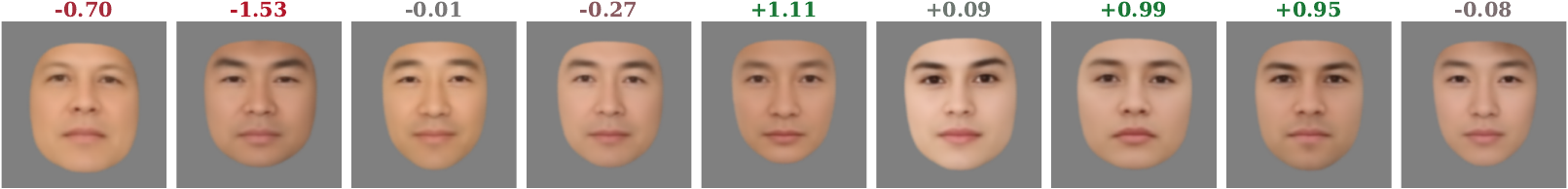}%
    
    \includegraphics[width=.9\linewidth]{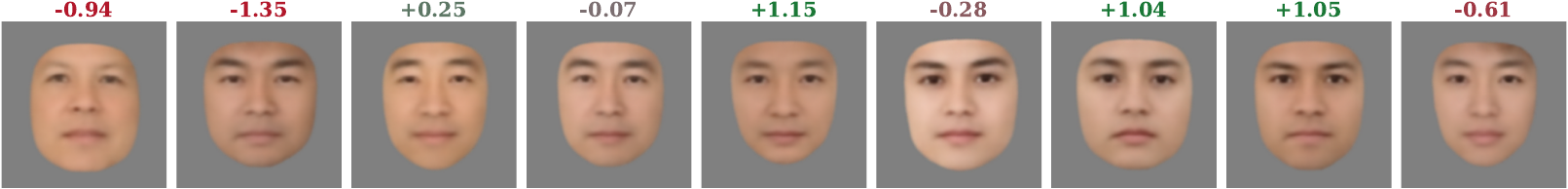}%
 }
\end{figure}

\newpage

\begin{figure}[!h]
\floatconts
  {fig:reconstructions}
  {\caption{
    Reconstructions of Black  Female (top) and Male (bottom) faces. Rows: (1) original CFD images, (2-5) Reconstructions from VAE trained on (2) FairFace, (3) FFHQ, (4) CelebA, (5) UTKFace. See Fig.~\ref{fig:examples} for details. 
  }}
  {%
    \includegraphics[width=.9\linewidth]{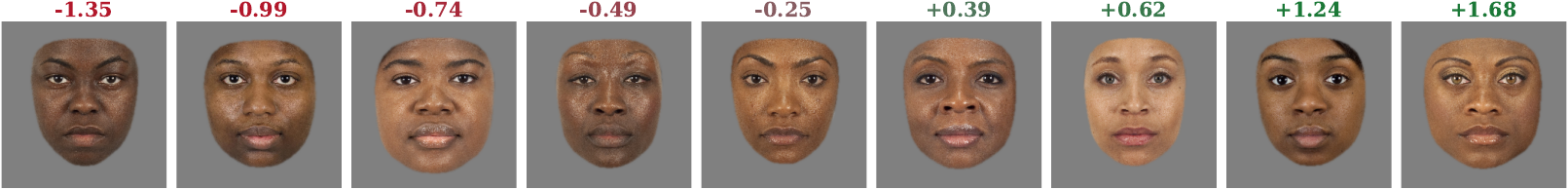}%
    
    \includegraphics[width=.9\linewidth]{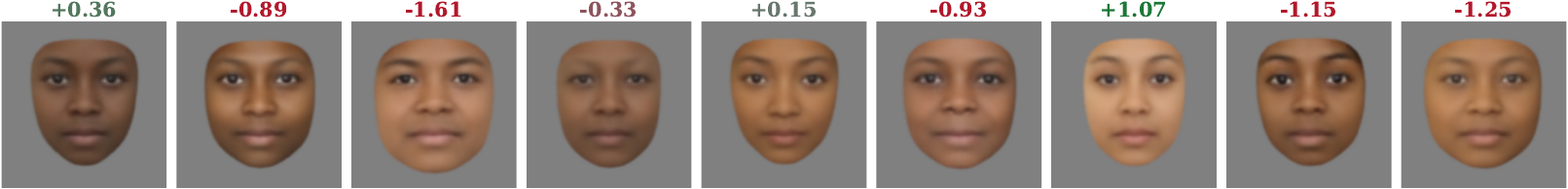}%
    
    \includegraphics[width=.9\linewidth]{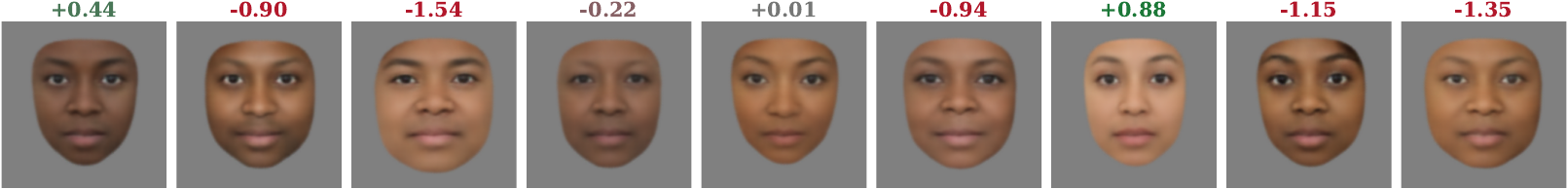}%
    
    \includegraphics[width=.9\linewidth]{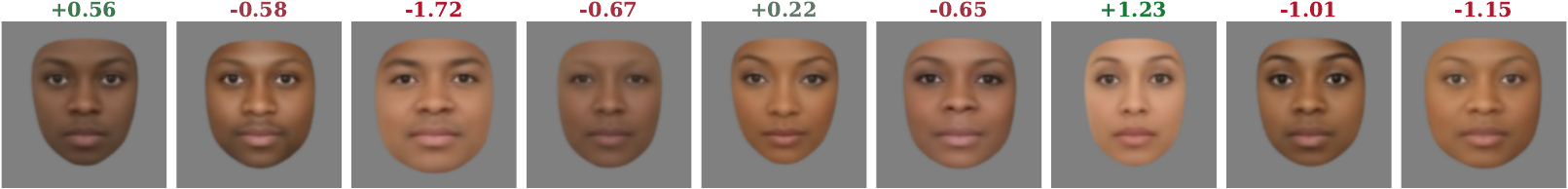}%
    
    \includegraphics[width=.9\linewidth]{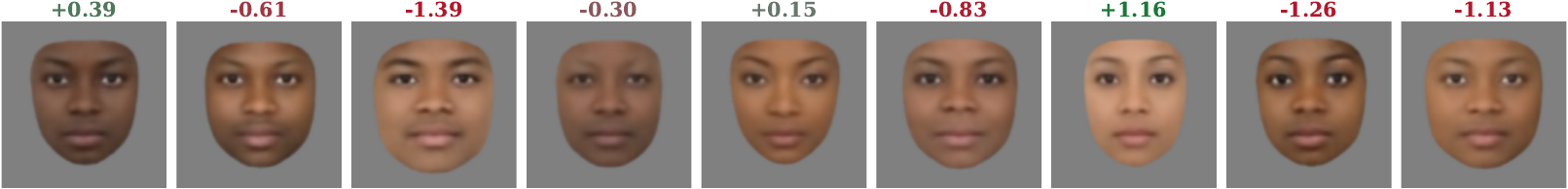}%

    \vspace{2em}

    \includegraphics[width=.9\linewidth]{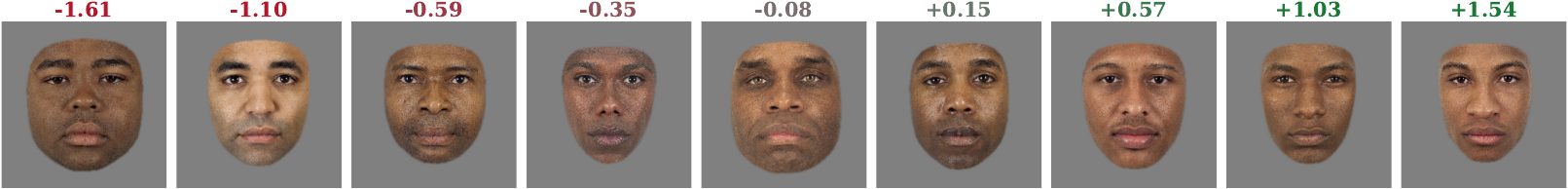}%
    
    \includegraphics[width=.9\linewidth]{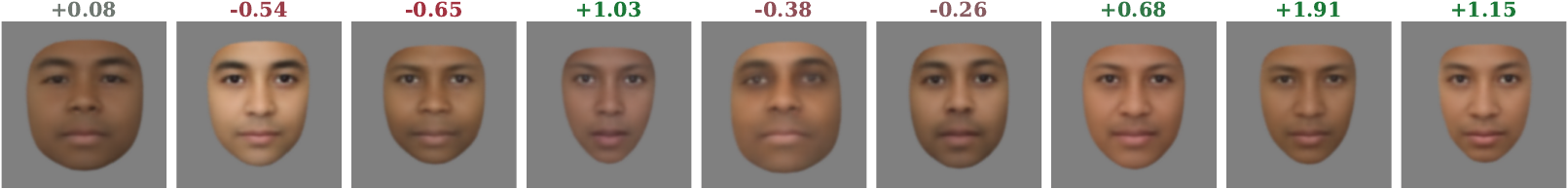}%
    
    \includegraphics[width=.9\linewidth]{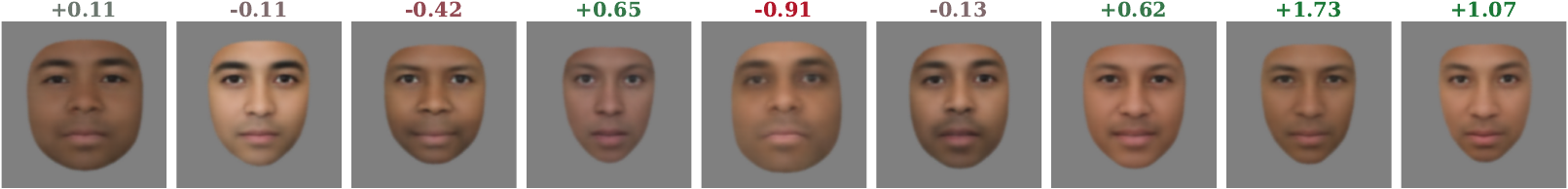}%
    
    \includegraphics[width=.9\linewidth]{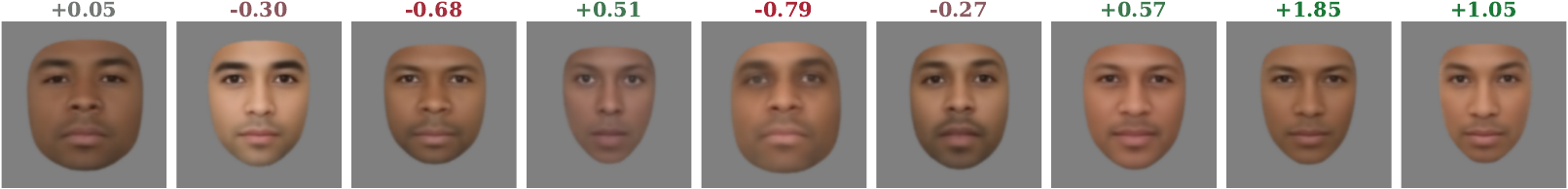}%
    
    \includegraphics[width=.9\linewidth]{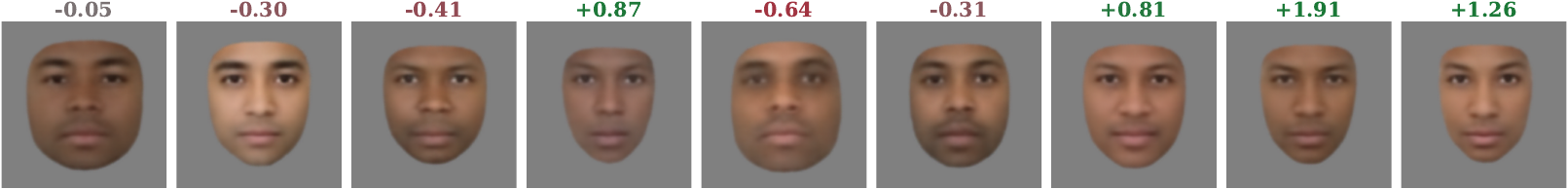}%
 }
\end{figure}

\newpage

\begin{figure}[!h]
\floatconts
  {fig:reconstructions}
  {\caption{
    Reconstructions of Latino  Female (top) and Male (bottom) faces. Rows: (1) original CFD images, (2-5) Reconstructions from VAE trained on (2) FairFace, (3) FFHQ, (4) CelebA, (5) UTKFace. See Fig.~\ref{fig:examples} for details.  
  }}
  {%
    \includegraphics[width=.9\linewidth]{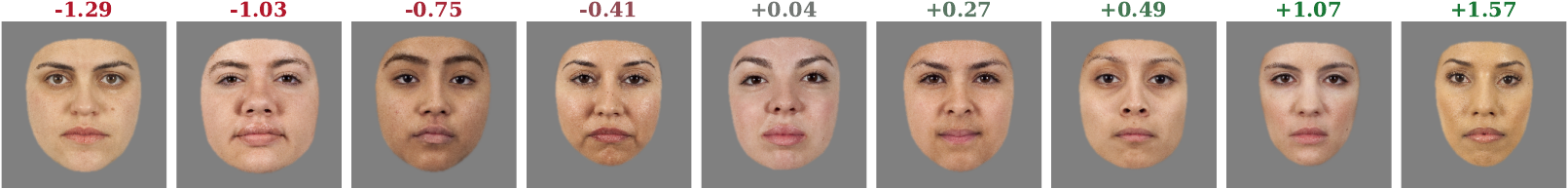}%
    
    \includegraphics[width=.9\linewidth]{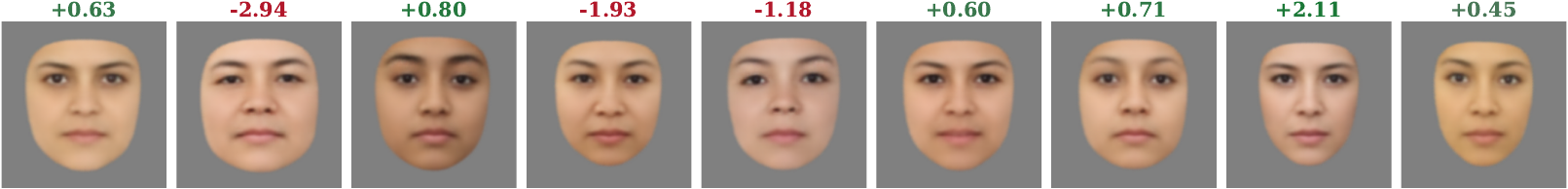}%
    
    \includegraphics[width=.9\linewidth]{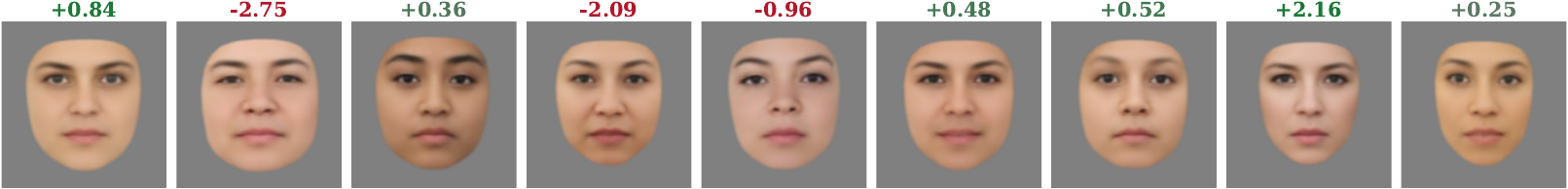}%
    
    \includegraphics[width=.9\linewidth]{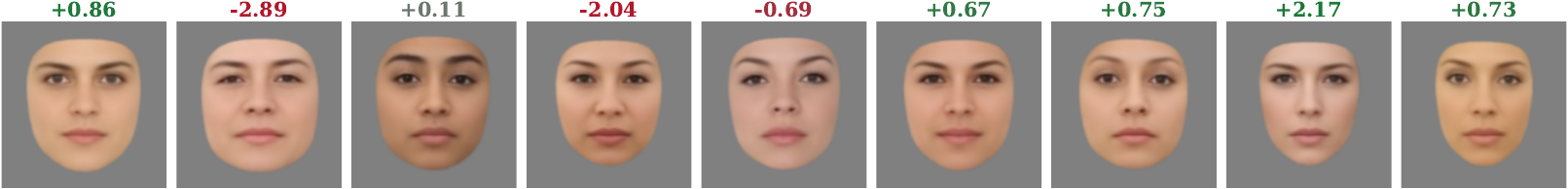}%
    
    \includegraphics[width=.9\linewidth]{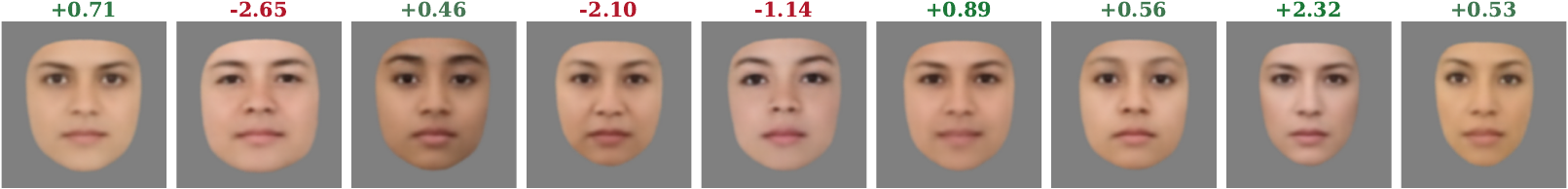}%

    \vspace{2em}

    \includegraphics[width=.9\linewidth]{FIG/reconstructions/LM/row_original.pdf}%
    
    \includegraphics[width=.9\linewidth]{FIG/reconstructions/LM/row_fairface.pdf}%
    
    \includegraphics[width=.9\linewidth]{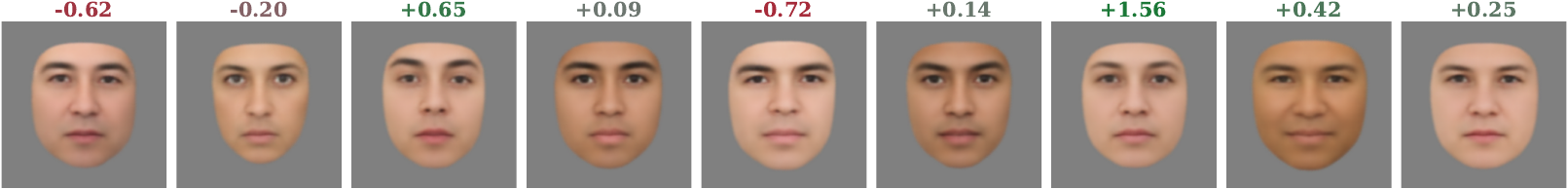}%
    
    \includegraphics[width=.9\linewidth]{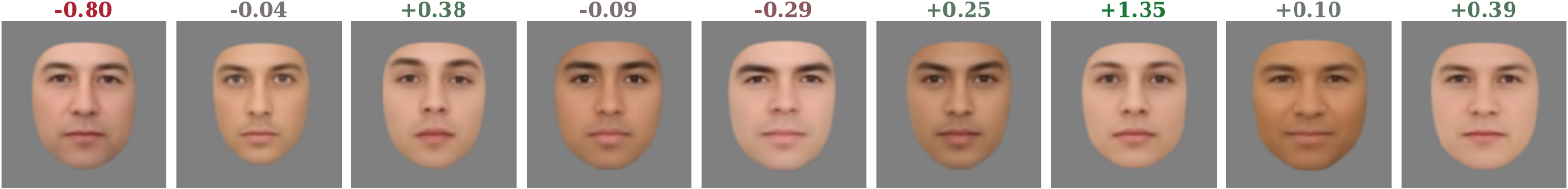}%
    
    \includegraphics[width=.9\linewidth]{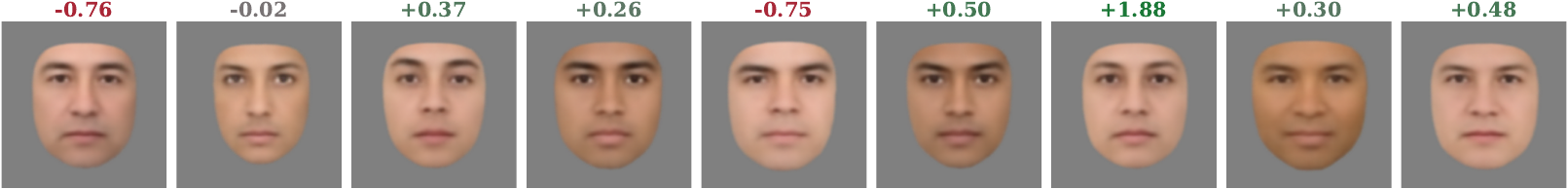}%
 }
\end{figure}

\newpage

\begin{figure}[!h]
\floatconts
  {fig:reconstructions}
  {\caption{
    Reconstructions of White  Female (top) and Male (bottom) faces. Rows: (1) original CFD images, (2-5) Reconstructions from VAE trained on (2) FairFace, (3) FFHQ, (4) CelebA, (5) UTKFace. See Fig.~\ref{fig:examples} for details.  
  }}
  {%
    \includegraphics[width=.9\linewidth]{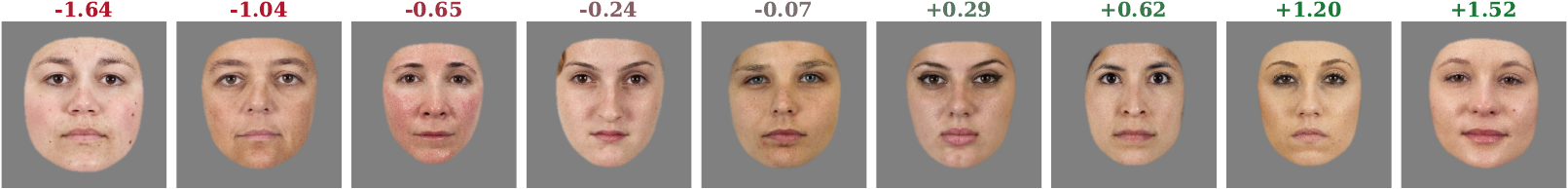}%
    
    \includegraphics[width=.9\linewidth]{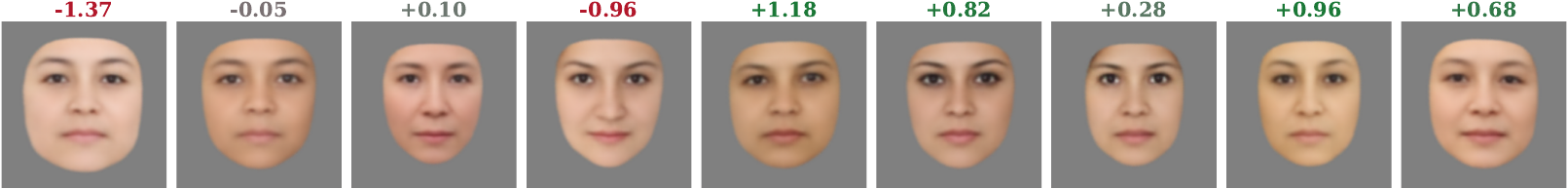}%
    
    \includegraphics[width=.9\linewidth]{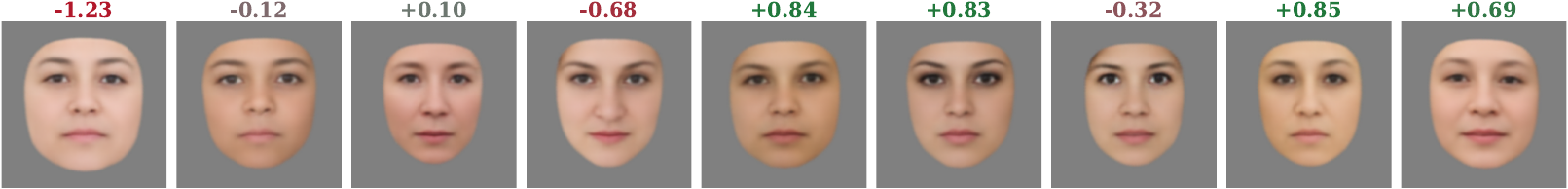}%
    
    \includegraphics[width=.9\linewidth]{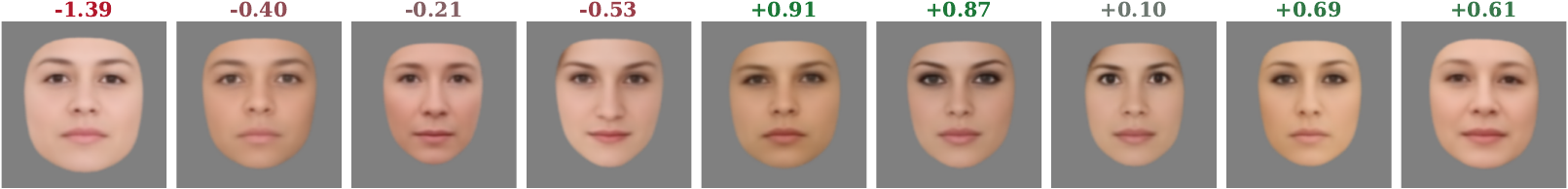}%
    
    \includegraphics[width=.9\linewidth]{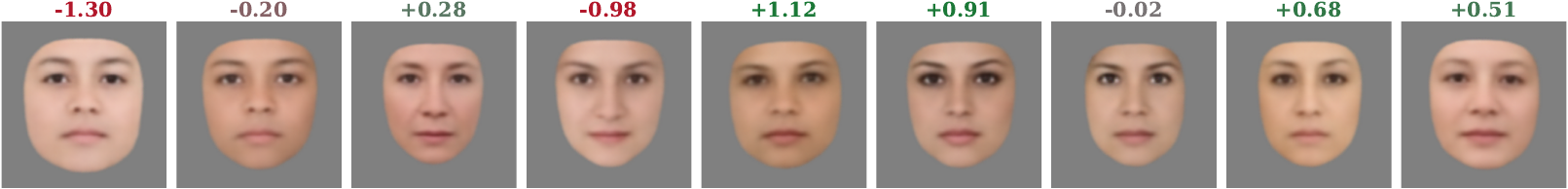}%

    \vspace{2em}

    \includegraphics[width=.9\linewidth]{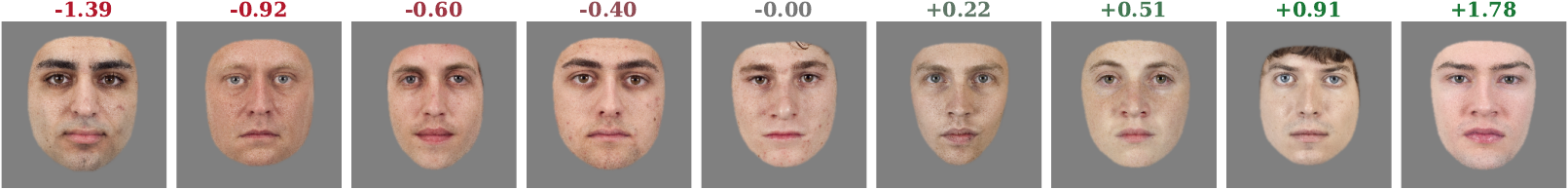}%
    
    \includegraphics[width=.9\linewidth]{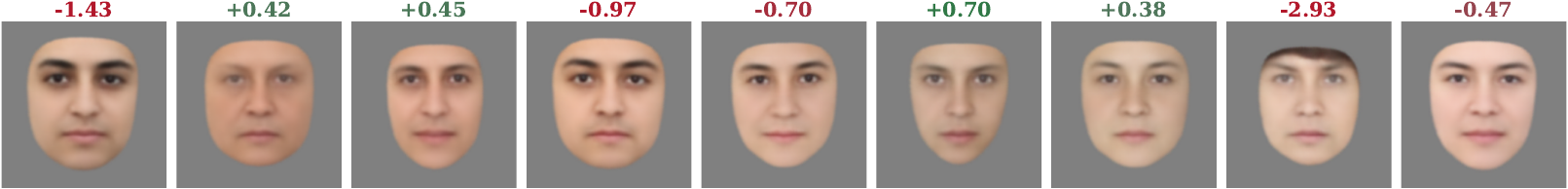}%
    
    \includegraphics[width=.9\linewidth]{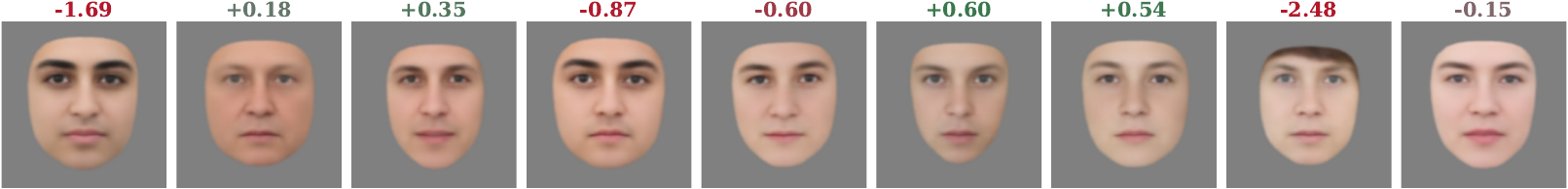}%
    
    \includegraphics[width=.9\linewidth]{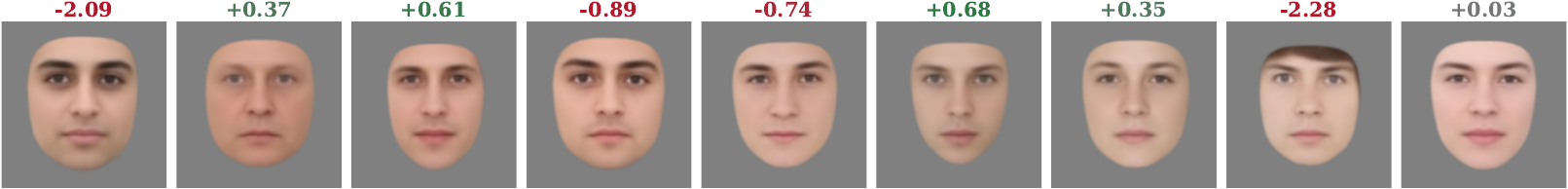}%
    
    \includegraphics[width=.9\linewidth]{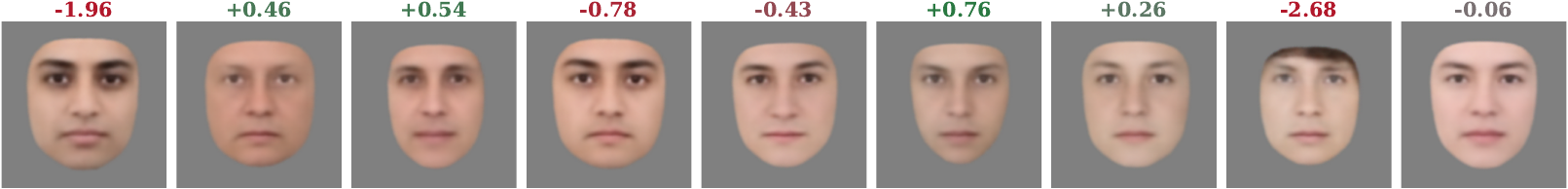}%
 }
\end{figure}

\end{document}